\documentclass[letterpaper, 10 pt, conference]{ieeeconf}  

\IEEEoverridecommandlockouts                              

\usepackage{amsmath} 
\usepackage{amssymb}  
\usepackage{graphicx}

\title{\LARGE \bf
ArtiTwinSplat: Interactable Digital Twin Reconstruction via Gaussian Splatting from RGB-D videos
}

\author{Pranjal Mishra$^{1}$, René Zurbrügg$^{1}$, Max Wilder-Smith$^{1}$, 
Marco Hutter$^{1}$, Marc Pollefeys$^{1,2}$,\\ Zuria Bauer$^{1}$, 
Hermann Blum$^{3}$%
\thanks{$^{1}$ETH Zürich, Zürich, Switzerland.}%
\thanks{$^{2}$Microsoft, Zürich, Switzerland.}%
\thanks{$^{3}$University of Bonn, Bonn, Germany.}%
}

\begin{document}

\maketitle
\thispagestyle{empty}
\pagestyle{empty}

\begin{abstract}

Deploying robots in unstructured real-world environments needs accurate, interactive models of the objects. Constructing these models at scale remains a critical bottleneck for robotic system integration. We present ArtiTwinSplat, a framework that automatically constructs articulated, photo-realistic digital twins of objects directly from RGB-D videos, requiring no CAD models, simulation assets, or manual annotations. Our method is built on 3D Gaussian Splatting that preserve geometric fidelity and photometric realism, coupled with an unsupervised articulation discovery pipeline that recovers part structure and joint kinematics from observed motion alone. With tracking and optimization stages our method provides stable, queryable digital twins that support real-time rendering, viewpoint control, and interactive manipulation. Unlike prior methods confined to simulation, ArtiTwinSplat operates directly on real-world observations and produces twins that are immediately usable by downstream robot planning and learning systems. This method offers a practical, scalable pathway toward digital twin construction, lowering the integration barrier for articulated object manipulation in embodied AI and human-robot collaboration contexts.

\end{abstract}

\begin{figure*}[!t]
    \centering
    \resizebox{\textwidth}{!}{\includegraphics{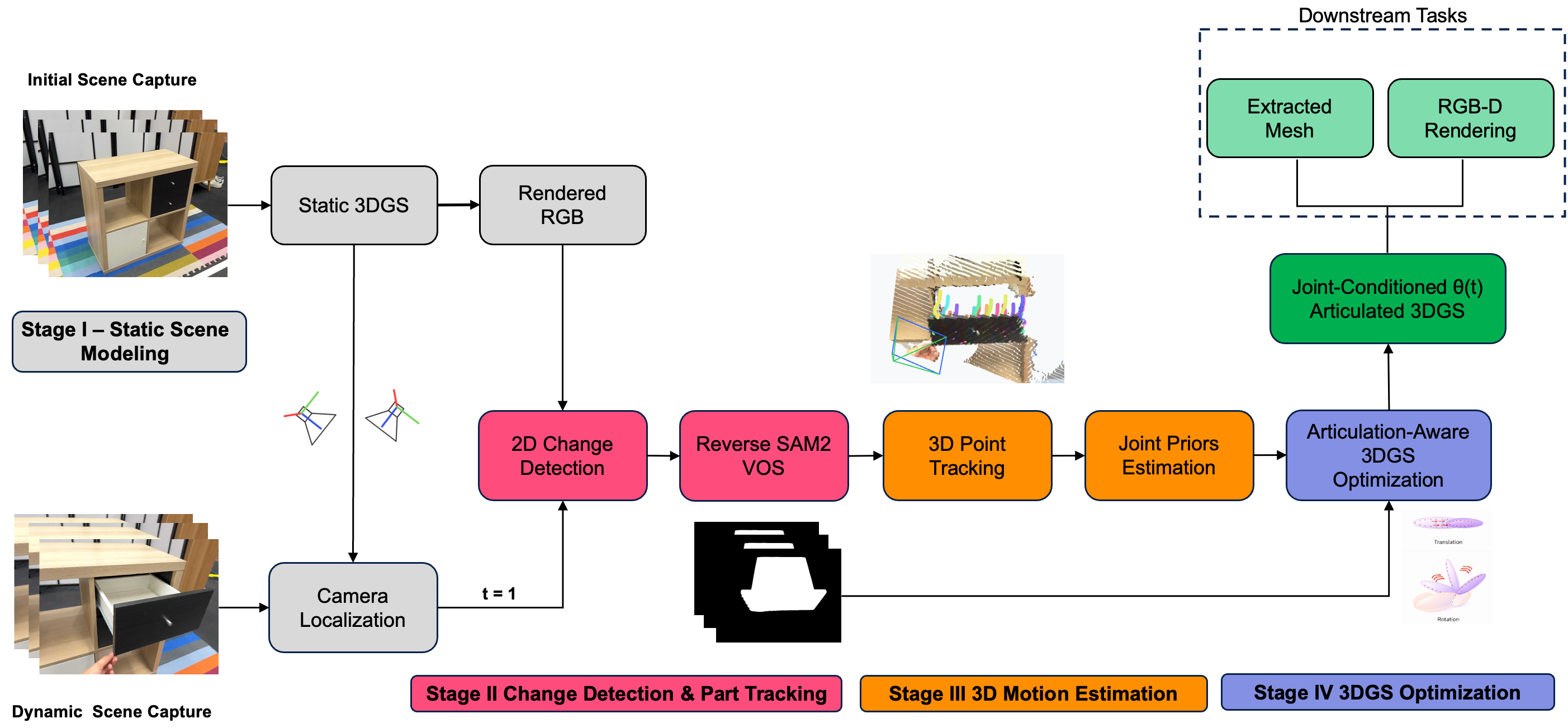}}
    \caption{\textbf{ArtiTwinSplat} pipeline for unsupervised articulated 3D reconstruction:
        (Stage I) A static pre-change sequence to train a canonical 3DGS model.
        (Stage II) A dynamic RGB-D capture is localized to the canonical model, and 2D appearance differences generate an initial change mask that seeds reverse SAM2 video object segmentation, giving dense per-frame object masks.
        (Stage III) Pixel correspondences are lifted to 3D for point tracking, and joint parameters: axis, pivot, and articulation variable $q(t)$ per frame are estimated via joint priors estimation.
        (Stage IV) These priors condition articulation-aware 3DGS optimization, producing a joint-conditioned articulated 3D representation.
    }
    \label{fig:pipeline}
\end{figure*}

\section{INTRODUCTION}

Articulated objects like drawers, doors, cabinets, and windows are central to everyday human-environment interactions and have become a key focus in computer vision and robotics \cite{c1, c3}. These objects appear in virtually every indoor environment, making their accurate perception and modeling essential for both scene understanding and physical manipulation. Accurate reconstruction of articulated objects is vital for a broad range of downstream applications, including scene understanding \cite{c4, c5, c6}, robotic interactions \cite{c7, c8}, and digital twin creation ~\cite{c9, c10}. High-fidelity digital twins that jointly capture appearance, geometry, and articulation enable robust sim-to-real transfer for embodied agents and support learning-based reasoning in interactive 3D environments.

Despite growing interest, articulated object reconstruction remains an open and difficult problem \cite{c2, c11}. The core difficulty lies in the tight coupling between geometry, appearance, and part-level motion: recovering structure and articulation jointly from dynamic observations is inherently ill-posed. Off-the-shelf reconstruction methods such as NeRF \cite{c12} and 3DGS \cite{c13} assume static scenes and do not generalize to articulated settings. Even under the simplifying assumption of rigid-body part motion, jointly estimating part geometry and kinematic parameters from a single moving camera remains highly underconstrained. Existing approaches that tackle this problem typically require manual part annotations, pre-segmented inputs, or controlled multi-view capture setups, these conditions rarely hold in practical robotic deployment \cite{c11, c16, c17, c14}.

In this work, we address this gap with ArtiTwinSplat, a fully automatic framework for reconstructing articulated objects from handheld RGB-D videos. Rather than relying on part labels or multi-view rigs, our approach discovers articulation structure directly from appearance and depth observations. We detect articulated regions through appearance-change analysis, enforce temporal consistency via reverse-time propagation, and recover physically meaningful joint parameters. These motion priors then condition a joint-aware Gaussian optimization, yielding an articulated 3DGS model that supports continuous and physically consistent articulation. Our main contributions are:
\begin{itemize}
\item \textbf{Automatic articulation discovery:} an annotation-free pipeline that detects articulated parts and recovers joint type, axis, pivot, and range of motion directly from RGB-D video.

\item \textbf{Joint-conditioned optimization:} a two-phase fine-tuning strategy that enforces physically consistent part motion while maintaining global scene appearance, producing stable articulated digital twins.

\item \textbf{Simulation-ready output:} direct export of reconstructed models as URDFs compatible with Isaac Sim, enabling a practical workflow from RGB-D captures to deployment-ready robotic assets.
\end{itemize}

\section{RELATED WORK}

\noindent\textbf{Articulated Object Reconstruction.}
Early reconstruction approaches relied on parametric or skeleton-based representations requiring prior knowledge of joint structures \cite{c1}. Learning-based methods such as Real2Code\cite{c14} and ArticulateAnything \cite{c15} infer articulation parameters from visual input, but depend on large annotated datasets and generalize poorly to novel real-world scenes. Methods like PARIS \cite{c11} jointly model geometry and motion by decomposing objects into parts, yet assume pre-segmented inputs or access to multiple discrete object states — conditions rarely satisfied in unconstrained dynamic captures. More recent 3DGS-based approaches including ArtGS \cite{c16} and SplArt \cite{c17} achieve high-quality articulation modeling but similarly require multi-view or multi-state observations and explicit part supervision.
On the other hand, 3DGS \cite{c13} has become the dominant representation for photorealistic reconstruction because of the rendering efficiency. Extensions such as Deformable 3DGS \cite{c18} and Shape-of-Motion~\cite{c19} adapt the framework to non-rigid and dynamic settings by tracking 3D point trajectories across frames. 

In contrast, ArtiTwinSplat operates on a RGB-D sequence without part labels, pre-segmented inputs, or discrete articulation states. By unifying appearance-based motion discovery, joint parameter estimation, and articulation-conditioned Gaussian optimization in a single pipeline, our approach bridges the gap between dynamic scene reconstruction and interactive articulated digital twin construction.

\section{METHOD}
\label{sec:method}

\noindent\textbf{Overview.}
ArtiTwinSplat is a four-stage pipeline that takes a static pre-change RGB-D sequence and a dynamic post-change RGB-D video as input, and produces an interactive, articulated 3DGS digital twin as output. The stages are: (i) canonical scene reconstruction, (ii) change detection and part tracking, (iii) joint parameter estimation, and (iv) articulation-conditioned Gaussian optimization. An overview is shown in Figure~\ref{fig:pipeline}.

\noindent\textbf{Canonical Scene Reconstruction.} We first reconstruct a static 3DGS model of the scene from the pre-change sequence using a standard SfM pipeline~\cite{c21} for camera pose estimation, which produces a sparse point cloud defining the canonical scene geometry. The resulting 3DGS model act as both the geometric and photometric reference for all subsequent stages, and provides the initialization for Gaussian subset partitioning during optimization. Post-change frames are localized against this canonical point cloud to establish a shared coordinate frame across both sequences.

\noindent\textbf{Change Detection and Part Tracking.} We detect articulation-induced changes by comparing rendered canonical views against post-change frames in both appearance and geometry spaces, following~\cite{c22}. Specifically, dense feature embeddings are used to localize regions of motion, producing a coarse change mask at the final post-change frame. To obtain temporally consistent part masks across the full dynamic sequence, including frames with occlusions or partial visibility, we run SAM2~\cite{c20} in semi-supervised video object segmentation mode. The process is initialized from the final change mask and propagated in reverse temporal order from $t = T$ back to $t = 0$. Since the articulated part is most clearly visible and least occluded at the final frame, initializing there and tracking backward gives cleaner and more temporally stable segmentations.

\noindent\textbf{Joint Parameter Estimation.} Using per-frame SAM2 masks, we sample query points within the articulated region and track them across the full RGB-D sequence using TAPIP3D~\cite{c23}, which produces dense 3D point trajectories $\mathbf{p}_i(t) \in \mathbb{R}^3$ for $i=1,\ldots,N$ and $t=0,\ldots,T$ in world coordinates by leveraging persistent 
3D correspondences across time. These spatio-temporal trajectories are then passed to a 4D RANSAC procedure~\cite{c24} that jointly fits revolute and prismatic joint hypotheses by sampling minimal trajectory sets, computing inlier consensus, and selecting the model with the highest inlier ratio and lowest residual. The estimator outputs the joint type, axis $\mathbf{a}$, pivot $\mathbf{c}$, per-frame articulation variable $q(t)$, and range of motion $[q_{\min}, q_{\max}]$, all without manual annotation, part labels, or category-specific priors. Raw estimates are further stabilized through outlier filtering and temporal smoothing before being passed to the optimization stage. 

\noindent\textbf{Articulation-Conditioned Gaussian Optimization.}
The recovered joint parameters condition a two-phase fine-tuning of the canonical 3DGS model, which is initialized from the static pre-change sequence. In \textbf{Phase~I}, the dynamic post-change RGB-D frames are used as supervision. Scene Gaussians are partitioned into background $\mathcal{G}_{\mathrm{bg}}$, interior $\mathcal{G}_{\mathrm{interior}}$, and moving part $\mathcal{G}_{\mathrm{part}}$ subsets, where only $\mathcal{G}_{\mathrm{interior}} \cup \mathcal{G}_{\mathrm{part}}$ are trainable. Part Gaussians are transformed at each timestep according to

\begin{equation}
\mathbf{x}'_i(t) =
\left\{
\begin{array}{ll}
\mathbf{R}_{\mathbf{a},\,q(t)}(\mathbf{x}_i - \mathbf{c}) + \mathbf{c},
& \mathrm{revolute},\\[4pt]
\mathbf{x}_i + q(t)\,\mathbf{a},
& \mathrm{prismatic},
\end{array}
\right.
\end{equation}

and optimized under photometric, depth, and joint consistency losses within the SAM2 mask region.

In \textbf{Phase~II}, both the pre-change and post-change frames are used jointly to refine global scene radiance. All Gaussian subsets are unfrozen and optimized under weak photometric supervision while the learned articulation parameters $\{\mathbf{a}, \mathbf{c}, q(t)\}$ remain fixed, ensuring consistent illumination and appearance continuity across the full scene.

The final model supports real-time rendering and interactive joint control at arbitrary configurations, and can be exported as a URDF for use in physics simulators such as Isaac Sim.

\section{RESULTS}
\label{sec:results}
\noindent\textbf{Setup.}
We evaluate ArtiTwinSplat on three real-world household scenes captured with a handheld Apple iPhone Pro using the Record3D application, which provides synchronized RGB frames, metric depth maps, and camera poses via ARKit. No special capture rig are used at any stage. Each scene consists of two recordings: a static pre-change sequence captured before any object interaction, and a dynamic post-change sequence in which the articulated object is manipulated by hand. With SfM-based refinement~\cite{c21} is applied to the pre-change sequence to establish the canonical coordinate frame. All three scenes feature IKEA Kallax furniture units and cover both revolute joints (hinged cabinet doors) and prismatic joints (drawers), representing the articulated object types most commonly encountered in domestic robotic deployment.

\begin{figure}[t]
\centering
\setlength{\tabcolsep}{1.5pt}
\renewcommand{\arraystretch}{0.8}

\begin{tabular}{l ccc}
 & \small GT (Real) & \small Render & \small Depth \\[3pt]

\rotatebox{90}{\small Scene 1a}\; &
  \includegraphics[width=0.30\linewidth]{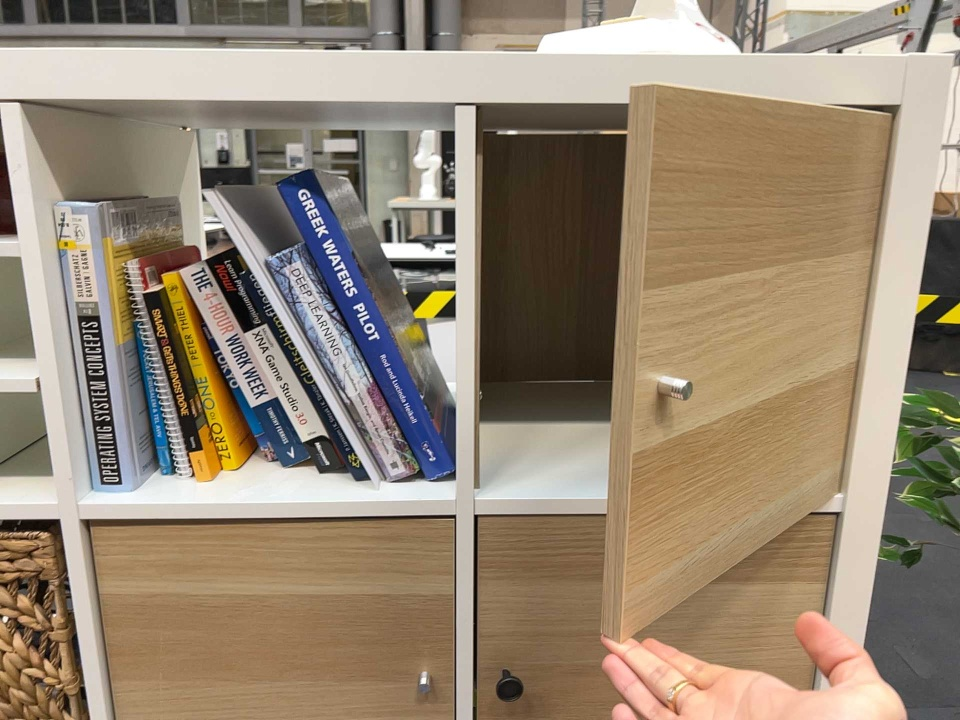} &
  \includegraphics[width=0.30\linewidth]{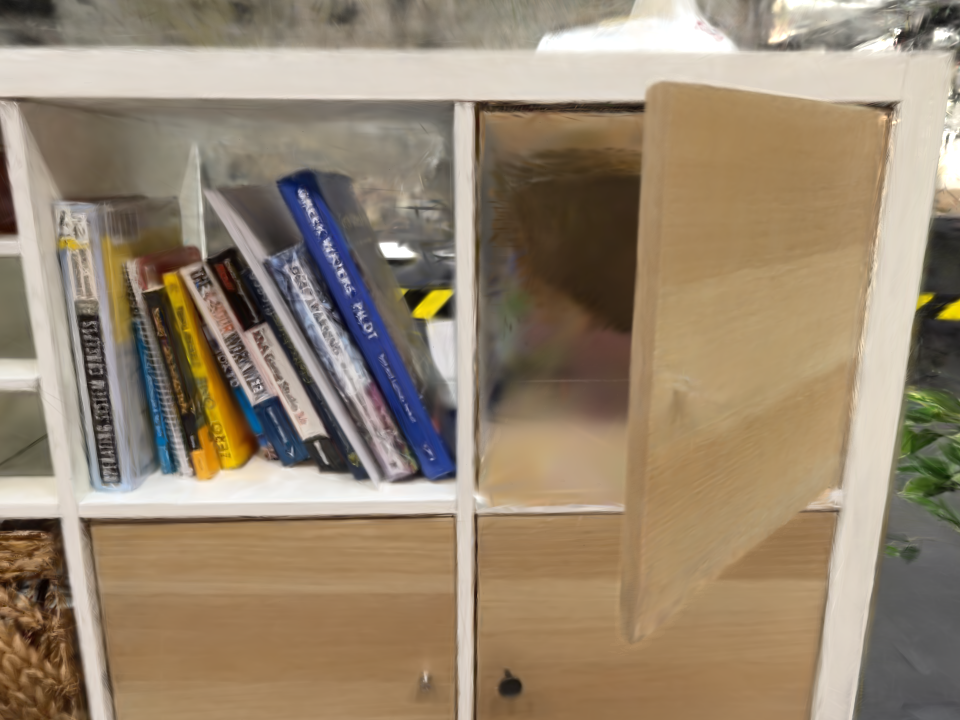} &
  \includegraphics[width=0.30\linewidth]{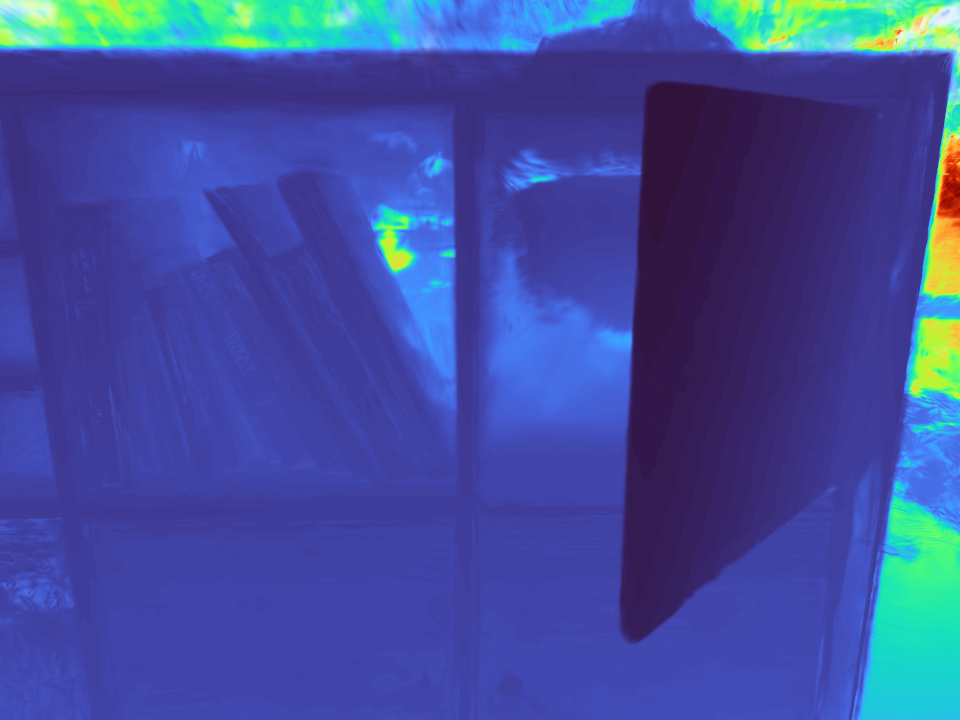} \\[3pt]

\rotatebox{90}{\small Scene 1b}\; &
  \includegraphics[width=0.30\linewidth]{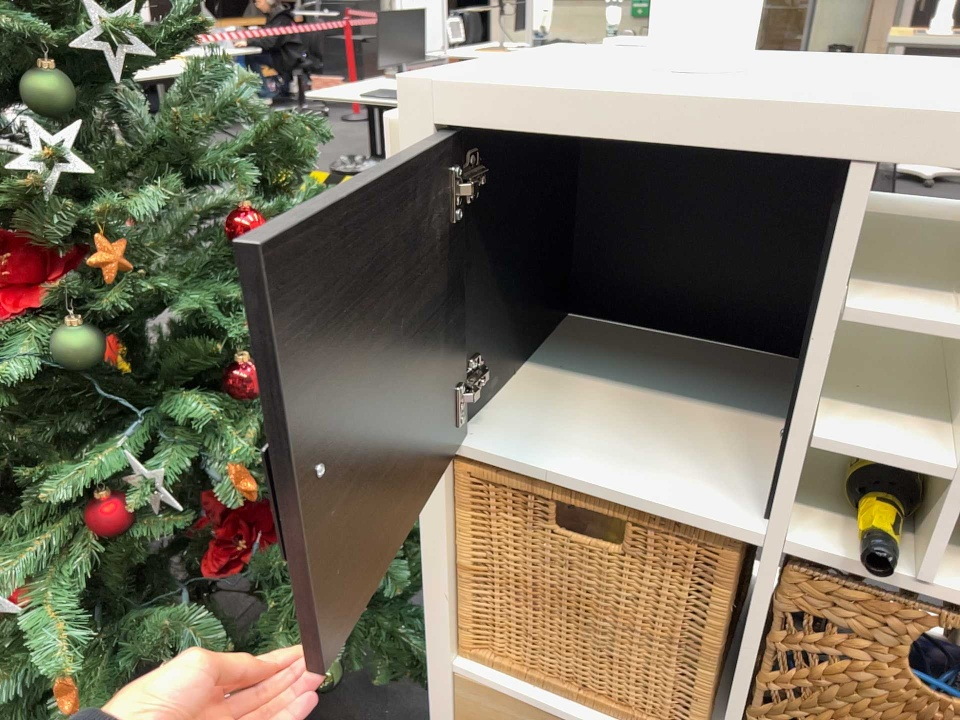} &
  \includegraphics[width=0.30\linewidth]{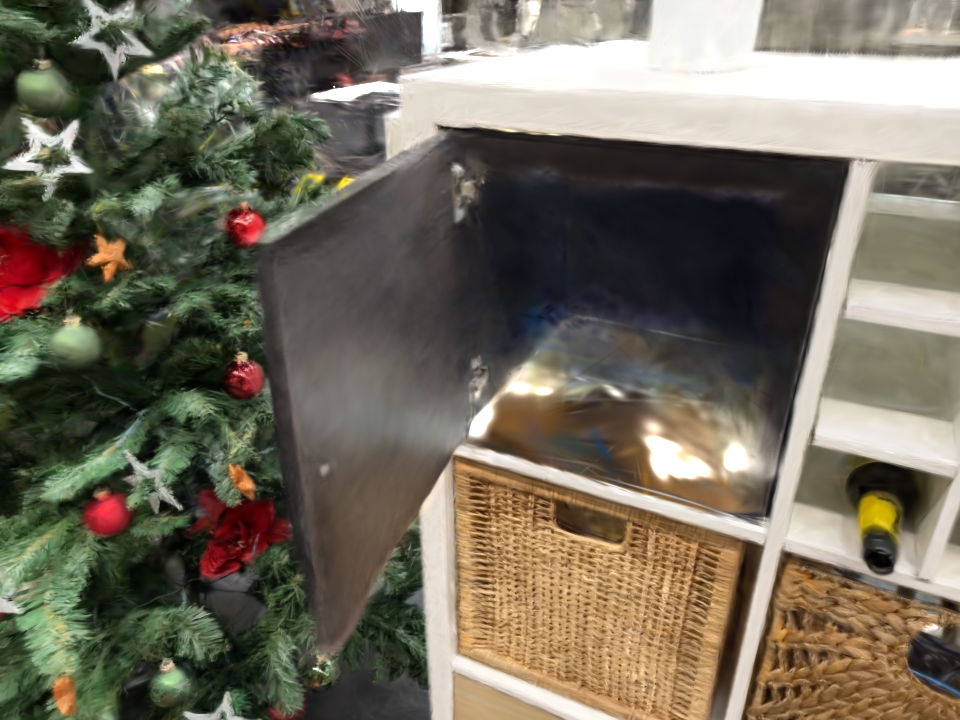} &
  \includegraphics[width=0.30\linewidth]{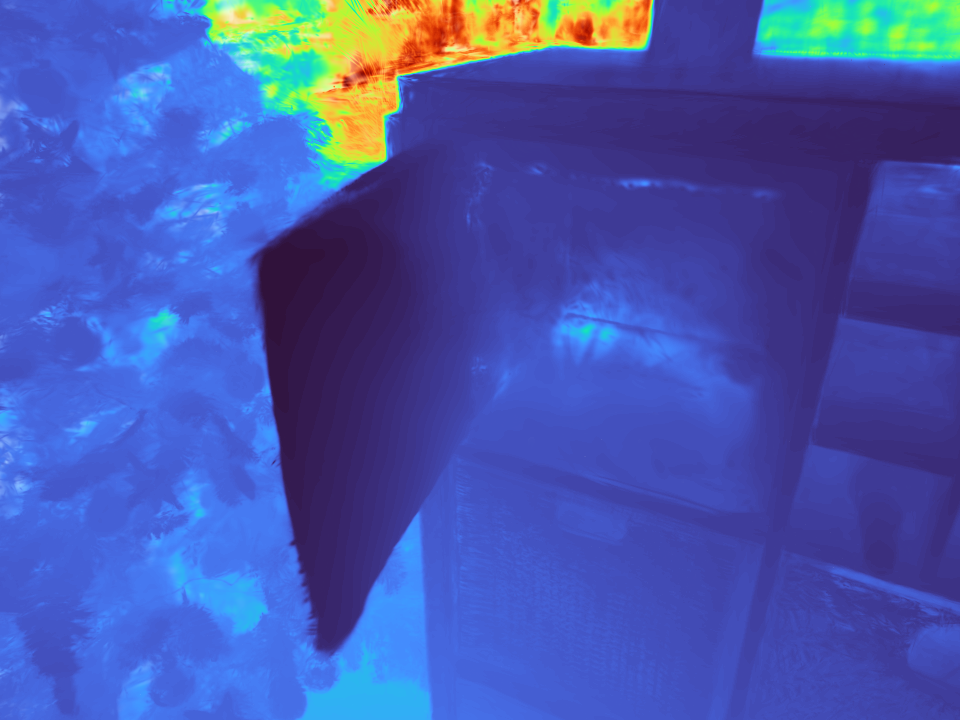} \\[6pt]

\rotatebox{90}{\small Scene 2a}\; &
  \includegraphics[width=0.30\linewidth]{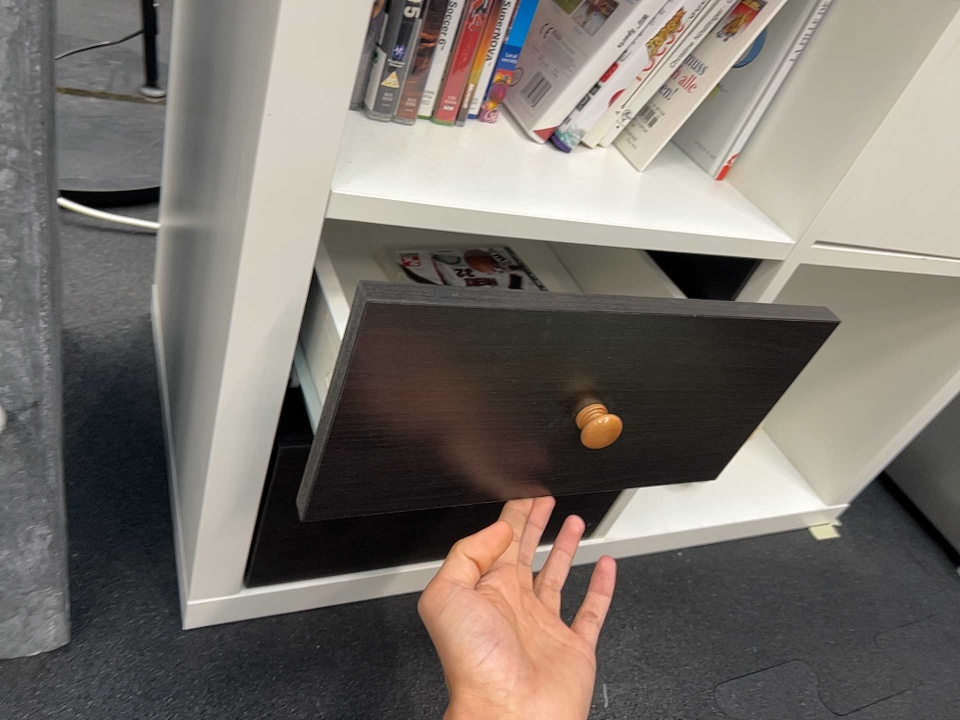} &
  \includegraphics[width=0.30\linewidth]{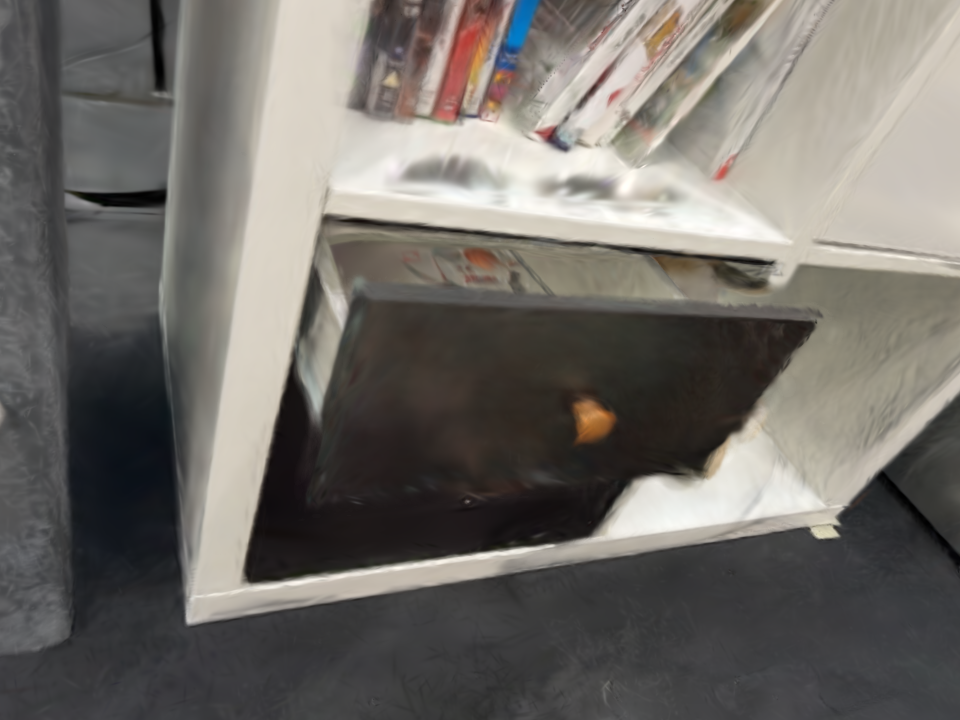} &
  \includegraphics[width=0.30\linewidth]{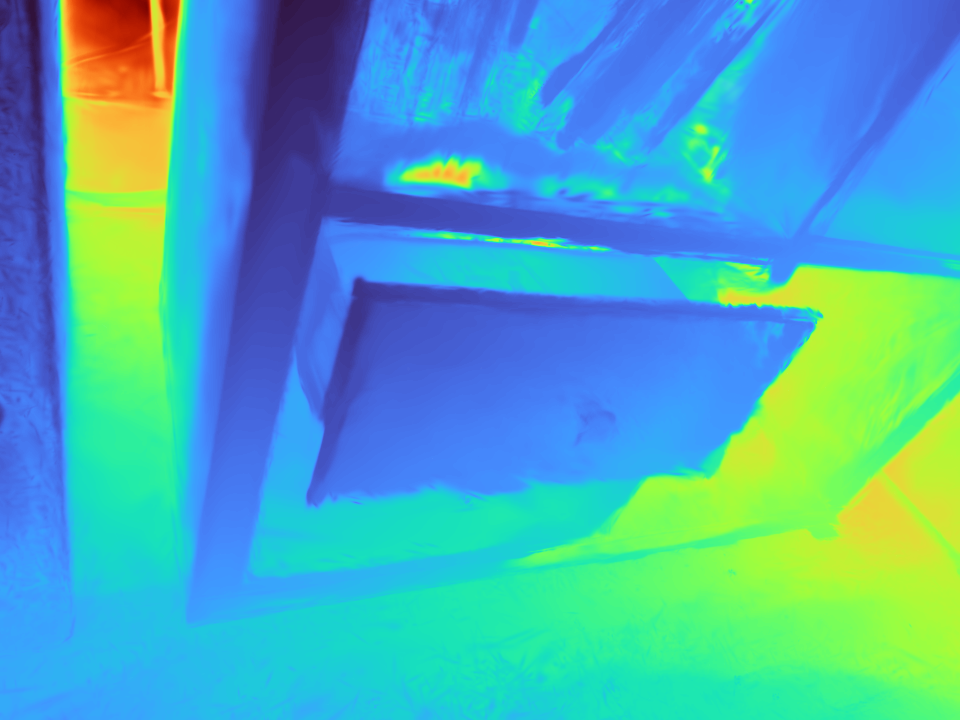} \\[3pt]

\rotatebox{90}{\small Scene 2b}\; &
  \includegraphics[width=0.30\linewidth]{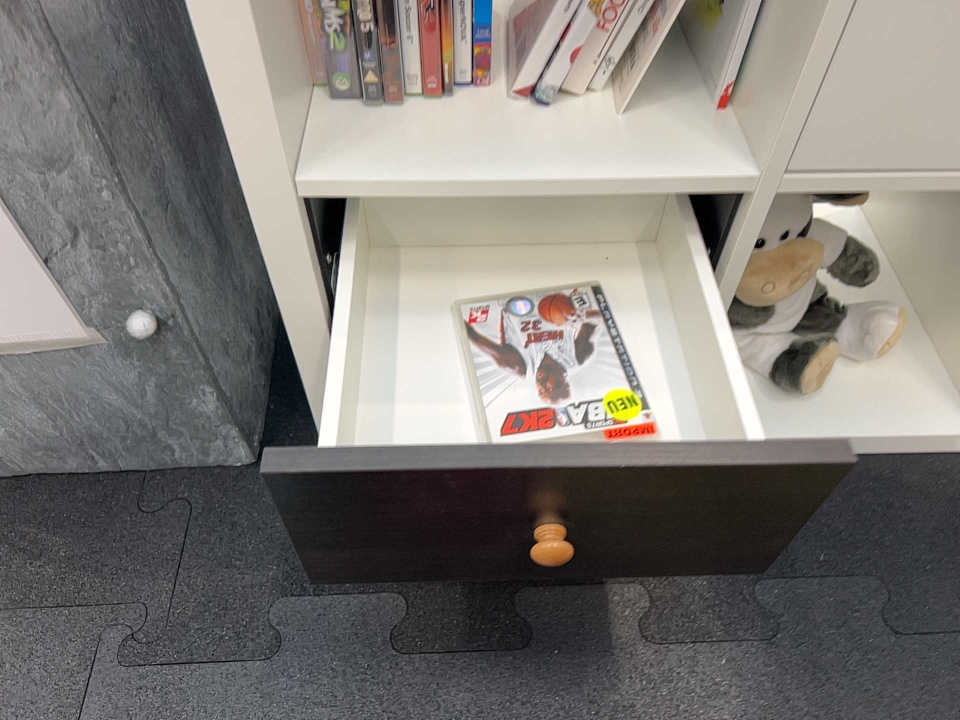} &
  \includegraphics[width=0.30\linewidth]{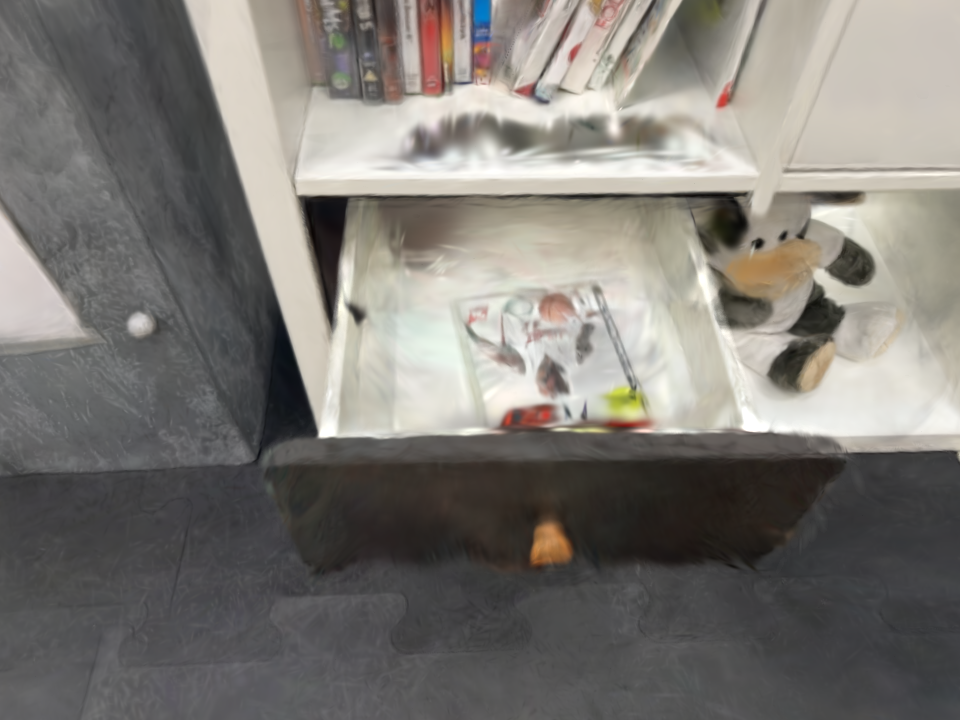} &
  \includegraphics[width=0.30\linewidth]{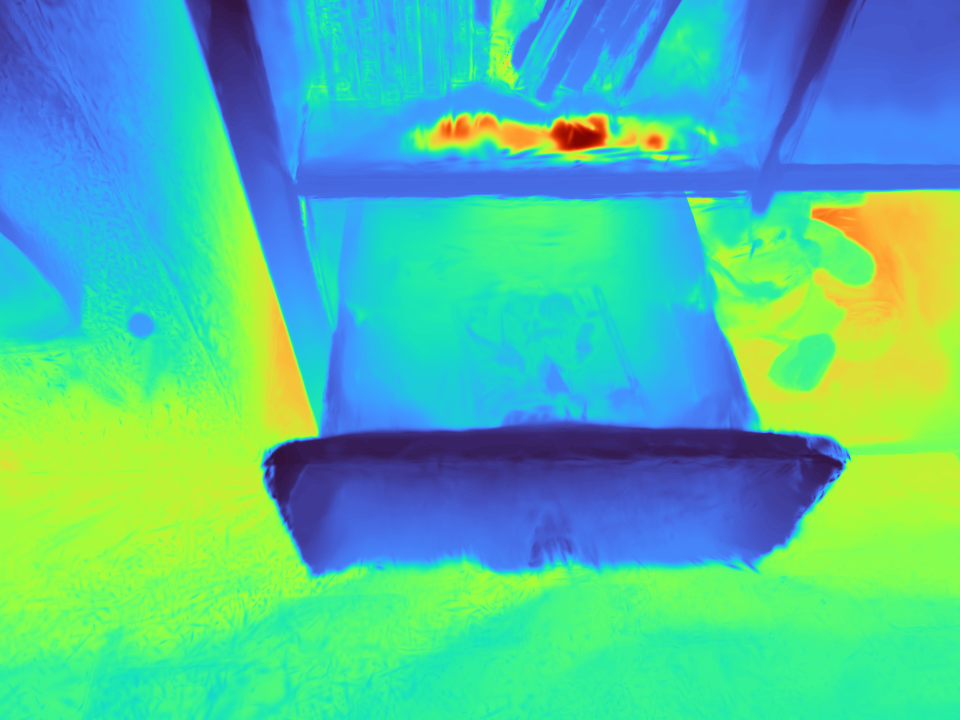} \\[6pt]

\rotatebox{90}{\small Scene 3a}\; &
  \includegraphics[width=0.30\linewidth]{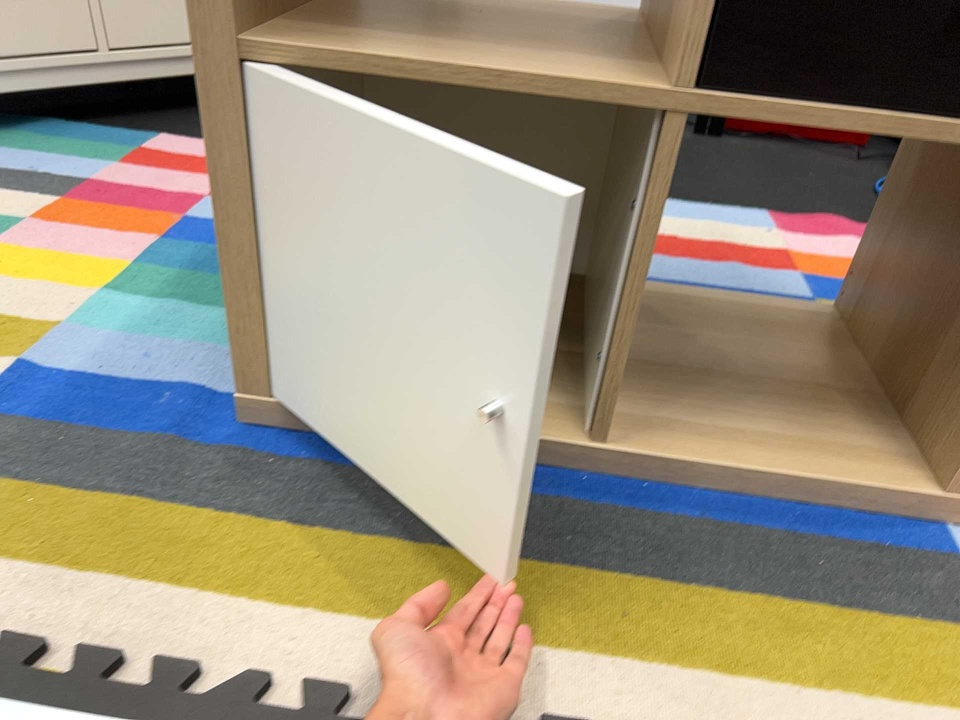} &
  \includegraphics[width=0.30\linewidth]{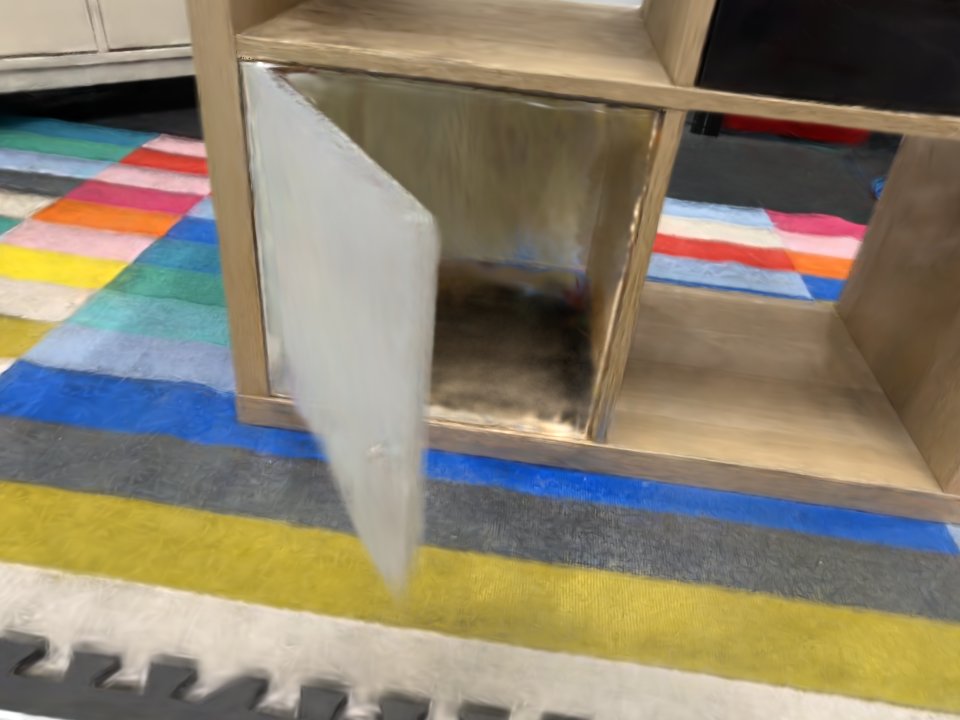} &
  \includegraphics[width=0.30\linewidth]{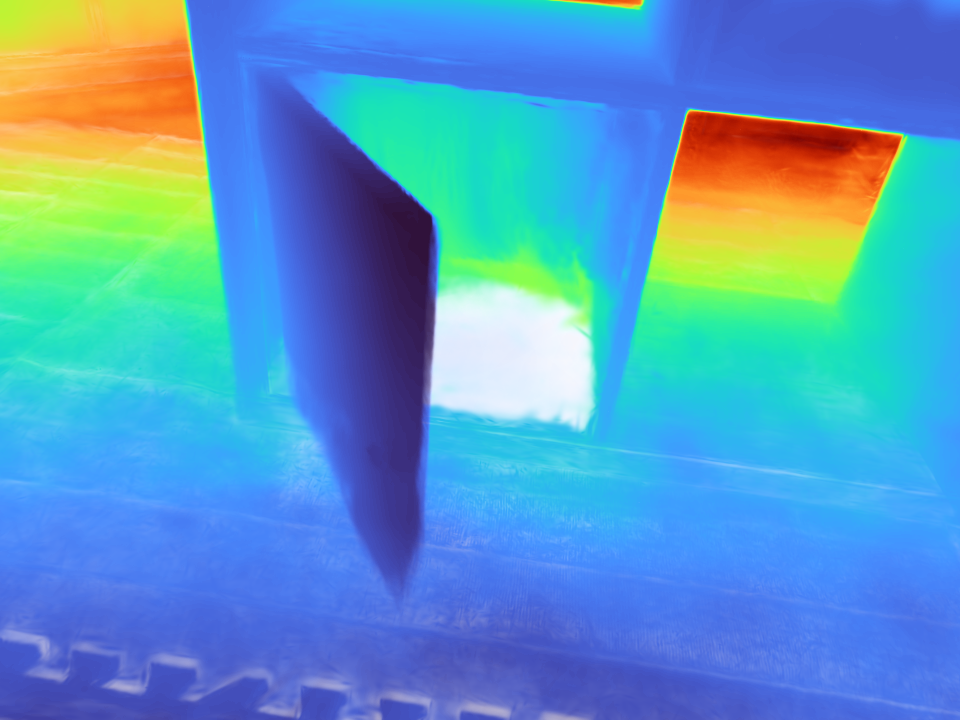} \\[3pt]

\rotatebox{90}{\small Scene 3b}\; &
  \includegraphics[width=0.30\linewidth]{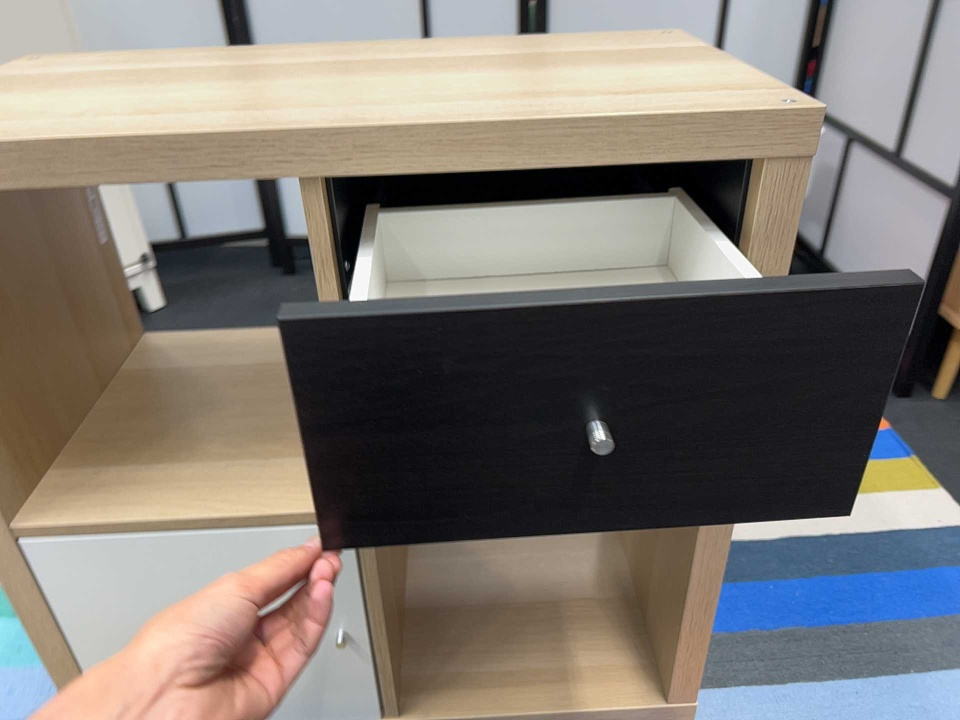} &
  \includegraphics[width=0.30\linewidth]{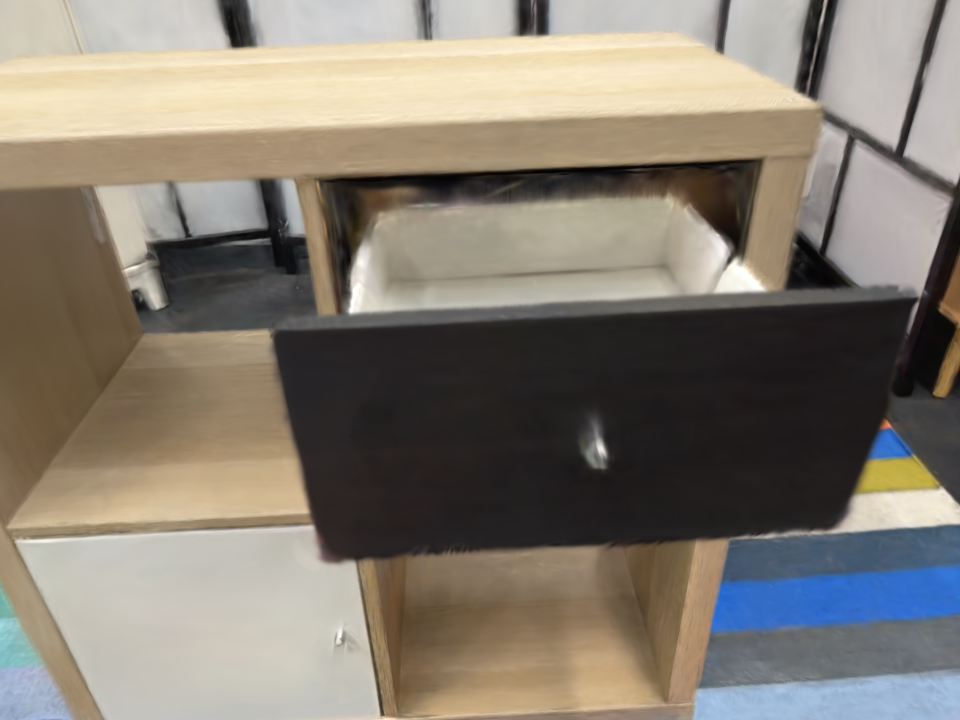} &
  \includegraphics[width=0.30\linewidth]{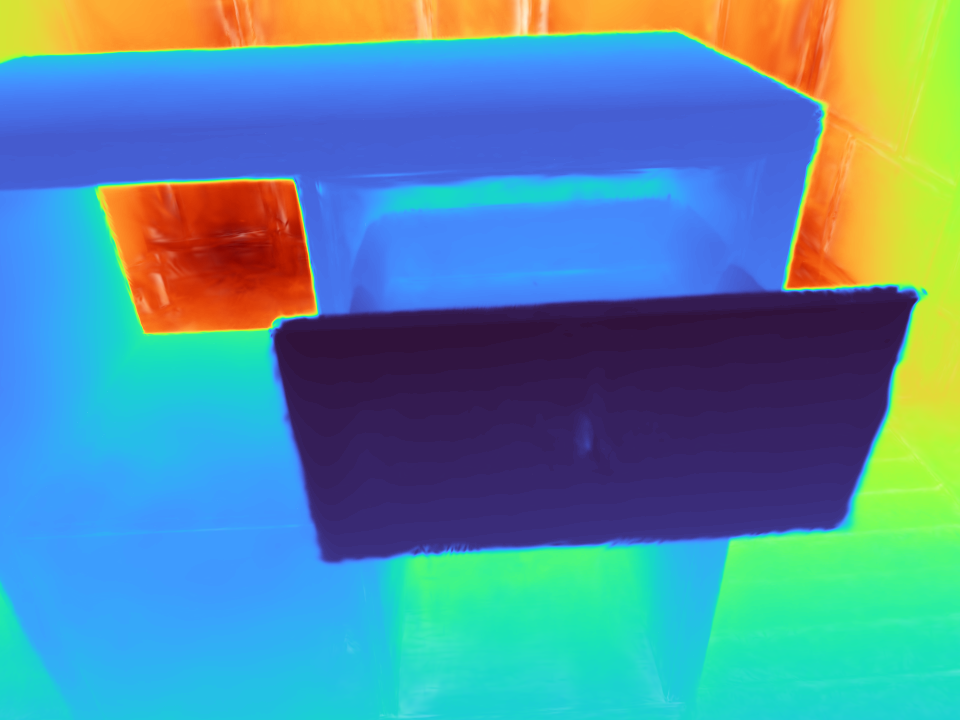} \\

\end{tabular}

\caption{Qualitative results across three real-world scenes at two 
articulation states each. For every state we show the real captured 
RGB frame (GT), our rendered RGB prediction, and the corresponding 
rendered depth map. ArtiTwinSplat faithfully reconstructs both 
appearance and geometry under revolute (Scene~1), prismatic (Scene~2), 
and mixed revolute+prismatic (Scene~3) articulation, directly from 
handheld RGB-D capture without manual annotation.}
\label{fig:realworld_renderings}
\end{figure}

\begin{figure}[!htb]
\centering
\setlength{\tabcolsep}{3pt}
\begin{tabular}{cc}
\small Real Capture & \small Isaac Sim Digital Twin \\[3pt]
\includegraphics[width=0.48\linewidth]{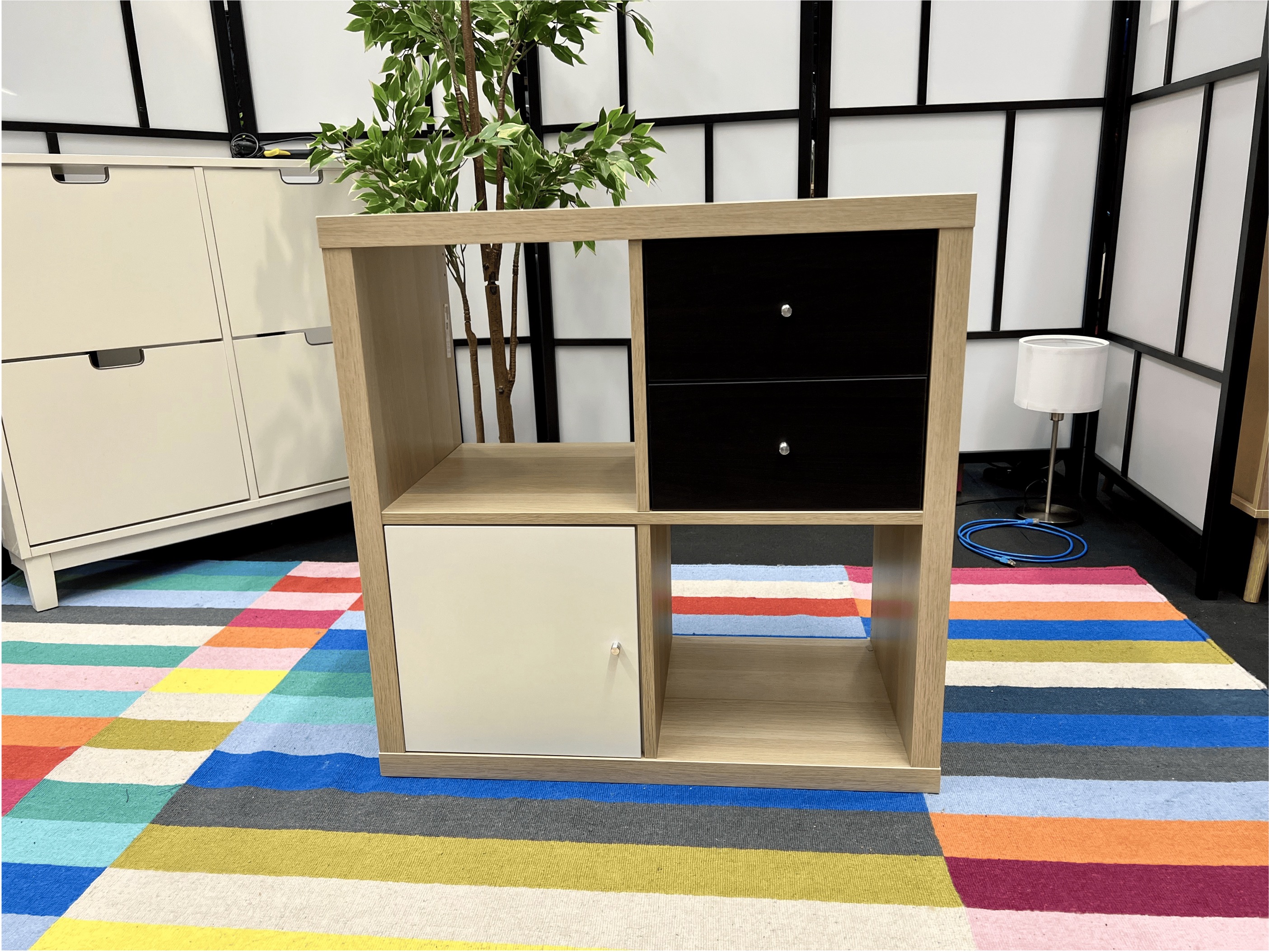} &
\includegraphics[width=0.48\linewidth]{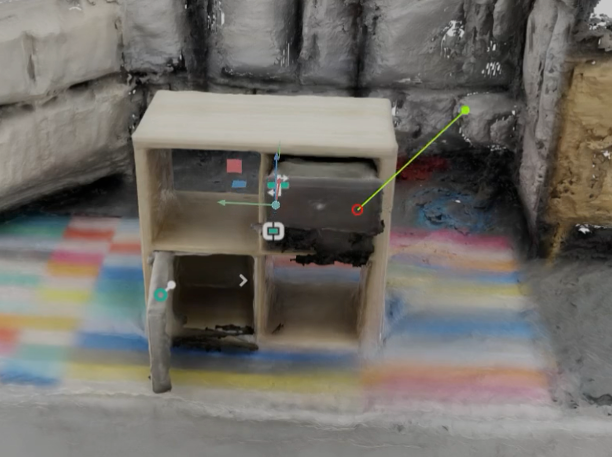}
\end{tabular}
\caption{From real-world capture to simulation-ready digital twin. 
\textit{Left:} A real household Kallax unit captured image. \textit{Right:} The reconstructed articulated digital twin 
imported into NVIDIA Isaac Sim, complete with collision meshes, revolute 
(hinged door) and prismatic (drawer) joint constraints, and URDF 
kinematics exported from ArtiTwinSplat}
\label{fig:isaacsim}
\end{figure}

\noindent\textbf{Results.}
Figure~\ref{fig:realworld_renderings} shows articulation-conditioned renderings at multiple joint states alongside real captured frames. Across all three scenes, our method reliably isolates moving components, recovers physically consistent joint parameters, and synthesizes photorealistic renderings under novel articulation configurations not seen during optimization. Prismatic reconstruction is particularly accurate due to consistent interior visibility throughout motion. Revolute joints are reconstructed faithfully at the exterior; occasional minor artifacts on interior surfaces stem from partial occlusion during rotation causing incomplete SAM2 mask coverage. The resulting models are directly exportable as URDFs compatible with Isaac Sim, providing simulation-ready articulated digital twins from a single casual RGB-D capture, as shown in Fig.~\ref{fig:isaacsim}.

\section{CONCLUSION}
\label{sec:conclusion}

We presented ArtiTwinSplat, a fully unsupervised framework for constructing  photo-realistic articulated digital twins from handheld RGB-D video.  By combining appearance-based change detection, SAM2-driven part tracking, joint estimation, and articulation-conditioned Gaussian optimization, our method recovers both geometry and kinematics of articulated objects without manual annotations, pre-segmented parts, or controlled capture setups. Qualitative results on real-world household scenes demonstrate photorealistic reconstruction across revolute and prismatic joint types, and the resulting models are directly exportable as simulation-ready URDFs for robotic manipulation and embodied AI applications.

\noindent\textbf{Limitations and Future Work.}
Our pipeline inherits limitations from upstream foundation models. SAM2 may over- or under-segment parts in regions with low texture or weak visual boundaries, and TAPIP3D can produce noisy correspondences under low texture or small motion, which degrades joint estimation. Interior surfaces that become visible only during revolute articulation remain challenging to reconstruct due to limited visibility and insufficient Gaussian supervision. Sensor depth noise further reduces geometric fidelity for thin structures. Future work will address these challenges through joint multi-part optimization, custom gaussian densification strategies, and tighter integration of 3D segmentation priors, moving toward a fully automated pipeline for scalable digital twin construction in unstructured environments.

\end{document}